\documentclass{article}

    \PassOptionsToPackage{numbers, compress}{natbib}

\usepackage[final]{neurips_2019}

\usepackage[utf8]{inputenc} 
\usepackage[T1]{fontenc}    
\usepackage{hyperref}       
\usepackage{url}            
\usepackage{booktabs}       
\usepackage{amsfonts}       
\usepackage{nicefrac}       
\usepackage{microtype}      

\usepackage{times}
\usepackage{soul}
\usepackage{url}
\usepackage[utf8]{inputenc}
\usepackage{graphicx}
\usepackage{amsmath}
\usepackage{booktabs}
\usepackage{algorithm}
\usepackage{algorithmic}
\usepackage{amsfonts,amssymb}
\usepackage{multirow}
\usepackage{subfig}
\usepackage{bm}
\usepackage[title]{appendix}
\usepackage{tabularx}
\newtheorem{theorem}{Theorem}[section]
\newtheorem{lemma}[theorem]{Lemma}

\DeclareMathOperator*{\argmax}{arg\,max}
\DeclareMathOperator*{\argmin}{arg\,min}
\DeclareMathOperator{\E}{\mathbb{E}}
\DeclareMathOperator{\PP}{\mathbb{P}}

\title{Guess First to Enable Better Compression and Adversarial Robustness}

\author{%
  Sicheng Zhu \\
  University of Virginia\\
  \texttt{sz6hw@virginia.edu} \\
  \And
  Bang An \\
  State University of New York at Buffalo \\
  \texttt{anbang@buffalo.edu} \\
  \And
  Shiyu Niu \\
  University of Electronic Science and Technology of China \\
  \texttt{niushiyuvip@gmail.com} 
}

\begin{document}

\maketitle

\begin{abstract}
Machine learning models are generally vulnerable to adversarial examples, which is in contrast to the robustness of humans. 
In this paper, we try to leverage one of the mechanisms in human recognition and propose a bio-inspired classification framework in which model inference is conditioned on label hypothesis. 
We provide a class of training objectives for this framework and an information bottleneck regularizer which utilizes the advantage that label information can be discarded during inference.
This framework enables better compression of the mutual information between inputs and latent representations without loss of learning capacity, at the cost of tractable inference complexity.
Better compression and elimination of label information further bring better adversarial robustness without loss of natural accuracy, which is demonstrated in the experiment.

\end{abstract}

\section{Introduction}

Current machine learning models are generally vulnerable to adversarial examples \cite{szegedy2013intriguing}, whereas in contrast, humans are arguably more robust (at least when given sufficient recognition time).
Recall that when an object is difficult to recognize, human tends to make hypotheses first (we slightly abuse the term 'hypothesis' through this paper to denote the 'guess') and then review evidence to test the hypotheses \cite{dicarlo2012does}.
This mechanism inspires us to extend the current classification framework.

Better compression of information theoretic measures between model input and latent representations has been shown beneficial in various aspects, such as better generalization \cite{xu2017information} and better adversarial robustness \cite{Alemi,pensia2019extracting}. Our proposed framework, by utilizing the bio-motivated mechanism, enables better compression in terms of mutual information without loss of learning capacity.

Concerns about the connection between the adversarial vulnerability of neural networks and the cross-entropy objective have been raised recently \cite{jacobsen2018excessive,chen2018complement}.
Motivated by the different empirical behaviors of different objectives observed by \citep{hjelm2018learning}, we further provide a class of training objectives for our framework which shows various empirical properties.

The commonly used variational information bottleneck \cite{Alemi} controls the mutual information by minimizing its variational upper bound, which can be solved analytically only when the representation distribution is parameterized as Gaussian. This restricts the use of information bottleneck to relatively high-level representations since low-level representations are unlikely to be Gaussian. We propose a new regularizer which, as a byproduct, does not have such restrictions.

To summarize, in this paper:
(1) We propose a novel classification framework in which model inference is conditioned on label hypothesis. This framework has several potential advantages to be utilized.
(2) Based on the proposed framework, we extend the commonly used cross-entropy loss to a new class of training objectives for classification tasks via f-divergence variational estimation. Objectives in this class have different properties which can be utilized to empirically stabilize training or improve adversarial robustness. 
(3) We propose an information bottleneck regularizer which aims to minimize the mutual information between inputs and latent representations. This regularizer takes advantage of the new framework, in which label information can be discarded during inference, to avoid certain shortcomings of commonly used variational information bottlenecks. We employ maximum mean discrepancy as a surrogate loss for efficient training. 
(4) We empirically test our framework by attack-based evaluations. Results show the adversarial robustness improvements.



\section{Proposed Method}
We propose a new classification framework in this section. First, we give problem formulation and the framework overview. Second, we derive the new class of training objectives. Then we propose the information bottleneck regularizer. An overall algorithm of this framework is given in Appendix \ref{algorithm}.

\subsection{Framework Overview}
\begin{figure}[t]
    \centering
    \includegraphics[width=0.6\textwidth]{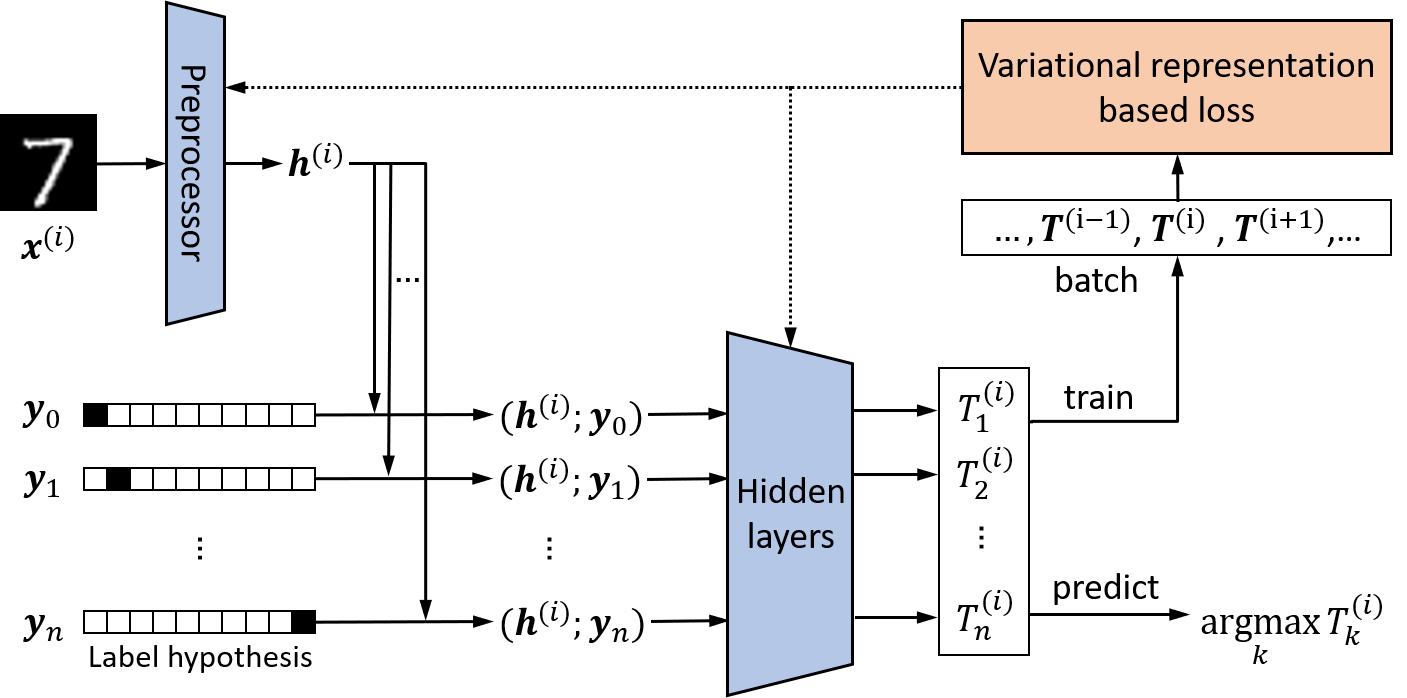}
    \caption{Architecture of the proposed framework. During inference, a sample $x^{(i)}$ is first fed into the preprocessing layers (e.g. primary convolutional layers). The obtained latent representation $h^{(i)}$ is then concatenated respectively with $n$ label embeddings $y_0, y_1,..., y_n$ (one-hot embedding in this example) to implement the label hypotheses mechanism. After propagation through the hidden layers, we get the outputs. For testing, the predicted label is the one with the largest output score. For training, the outputs of all samples within a batch will be put together to compute the proposed loss.}
    \label{fig:p1}
\end{figure}

The classification process of a typical feedforward neural network model can be viewed as a two-step procedure: (1) Map the input sample $x$ into latent representation $h$. (2) Embed label hypotheses into matrix $\bm{W}$ and project $h$ into the row space of $\bm{W}$ by $y=\bm{W}h$ (bias is included in the augmented form of $\bm{W}$), then choose the label hypothesis with the largest logit as the prediction. 
Inspired by the human recognition mechanism, we advance the hypothesis making procedure into the early stage of model inference. This results in an architecture illustrated in Figure \ref{fig:p1}. 

To be more specific, as in Figure \ref{fig:p1}, we leave some basic feature extraction layers unconditioned to reduce the computational complexity of inference. Note that for implementing the label hypotheses mechanism, there are multiple alternative ways to incorporate label information besides concatenation (e.g. embed labels to get projection matrices and then project $h$ into those matrices to get transformed vector). We use concatenation here to make the framework scalable when dealing with high dimensional data or large label categories.

\subsection{Classification Objectives}
The conventional neural network classification model can be interpreted as modeling of the conditional distribution $P(Y|X;\theta)$, where $\theta$ is the model parameter which is estimated using maximum likelihood principle when the training loss is cross-entropy. However, recent work shows the connection between the adversarial vulnerability of neural networks and the cross-entropy objective \cite{jacobsen2018excessive,chen2018complement}. Note that the probability density ratio satisfies $\frac{p(x,y)}{p(x)p(y)}=\frac{p(y|x)}{p(y)}$, so modeling the density ratio is equivalent to modeling conditional distribution when the marginal distribution of $Y$ is fixed. Therefore, we can model the conditional distribution and estimate the model parameter in an alternative way.

In general, we propose to use neural networks to parameterize a function $T(x,y)$, and model the conditional distribution by training $T(x,y)$ to be an approximation of the density ratio $\frac{p(x,y)}{p(x)p(y)}$. We show that $T(x,y)$ can be learned by solving an optimization problem that tightens the variational lower bound of the f-divergence between two fixed distributions.

We denote the joint and marginal distribution of two random variables $X$, $Y$ as $\mathbb{J}$ and $\mathbb{M}$, then the f-divergence between $\mathbb{J}$ and $\mathbb{M}$ has the form: 
$D_{f} (\mathbb{J} || \mathbb{M}) = \iint p(x,y) f\left(\frac{p(x,y)}{p(x)p(y)}\right)dxdy\label{eqn:df}$
where $f:\mathbb{R}\rightarrow \mathbb{R}$ is convex and lower semi-continuous function. There are multiple ways of leveraging a functional $T$ to represent $D_{f} (\mathbb{J} || \mathbb{M})$ by its variational lower bound. When the lower bound is tight, the optimal $T$ is a function of density ratio $\frac{p(x,y)}{p(x)p(y)}$ in closed-form. Therefore, we can parameterize $T$ with neural networks, optimize it to tighten the lower bound, and we arrive at an approximation of density ratio for further classification task.

Since there are multiple ways of representing f-divergence by its variational lower bound, we focus on three ways in this paper: (1) the general convex conjugate representation for f-divergences \cite{nguyen2010estimating,Nowozin}. (2) The Donsker-Varadhan representation for KL-divergence \cite{Donsker,belghazi2018mine}. (3) The energy-based representation for KL-divergence \cite{Poole}. It is worth noting that we are actually estimating the mutual information between the distributions of samples and labels when f-divergence is instantiated as KL-divergence. We apply the above three ways of variational representation to derive the objective. The first one is as follow and the other two are included in Appendix \ref{objectives}.

\textbf{The convex conjugate representation}. Utilizing the convex conjugate function $f^*(t)=\sup_{u \in \mathbb{R}}\{ut-f(u)\}$ of $f$, the f-divergence can be represented in a dual form followed with a lower bound:
\begin{align}
D_{f} (\mathbb{J} || \mathbb{M}) & \geq \sup_{T \in \mathcal{T}} \mathbb{E}_{\mathbb{J}}[T(x,y)] - \mathbb{E}_{\mathbb{M}}[f^*(T(x,y))] \label{eqn:cc}
\end{align}
where $\mathcal{T}$ is any class of functions $T:\Omega \rightarrow \mathbb{R}$. The bound is tight for $T^*(x,y)=f'(\frac{p(x,y)}{p(x)p(y)})$
where $f'$ denotes the derivative of $f$ (\cite{Nowozin}). As a result, the density ratio can be inversely solved from the optimal $T$. In many choices of f-divergence, the optimal $T$ is a monotonic function of density ratio. For example, JS-divergence has $T^*(x,y)=1+\text{log}\frac{2p(x, y)}{p(x,y)+p(x)p(y)}$. In this case,  $T^*(x,y)$ can be directly used as a score function in classification when the marginal distribution of labels is uniform. Note that the range of $T$ has to be limited to meet the domain constraint of $f^*$, this can be implemented by choosing proper activation function in the last layer of  $T$.  We use JS-divergence's representation in experiment which has the form:
\begin{align}
D_{\text{JS}} (\mathbb{J} || \mathbb{M}) 
\geq \sup_{T \in \mathcal{T}} \mathbb{E}_{\mathbb{J}}[\text{log} \sigma (T(x,y))] -  \mathbb{E}_{\mathbb{M}}[\text{log}(1- \sigma (T(x,y)))]\label{eqn:js}
\end{align}
where $\sigma$ is the sigmoid activation in the output layer. The other two objective representation forms are given in the appendix, together with the connection with the cross-entropy loss.

To use the variational representation in Equation \ref{eqn:js}, \ref{eqn:dv} and \ref{eqn:eb} as classification objective, we parameterize $T$ with neural networks and optimize the parameter to maximize the right hand side terms. We use true sample-label pairs within each batch to estimate the expectation over joint distribution $\mathbb{J}$. To estimate the marginal distribution $\mathbb{M}$, we generate marginal sample-label pairs by sampling $Y$ uniformly in all categories and pair them broadcastly with all samples $X$ within each batch. After the training process, we do predictions by finding $y^*$ with the largest output $T$: $y^* = \argmax_{y}T (x, y)$ since the marginal distribution of $Y$ in our experiment is uniform. 

When used as classification objectives, different choices of f-divergence representations offer different empirical benefits (e.g. training stability, adversarial robustness) in our experiment. 

\subsection{Regularization by Label Information Elimination}

Considering that adversarial perturbations are crafted under the guidance to mislead the label classification, we expect the distribution of sample representations with minimal label information to be invariant under the input perturbation and thus increase robustness against label adversarial attacks. Therefore, we propose to constrain the mutual information between labels and sample representations. Note that this may be too strong a regularization for a conventional classification model, since sample representations are meant to carry label information to the output layer and take inner product with label embeddings to pick the true label. While in our framework, label information can be eliminated or compressed once the label hypothesis is made. In experiments, we found overall significant robustness improvement after the regularizer is added.
  
To minimize mutual information, we first give the expression for mutual information between sample representation $H$ and ground-truth label $\tilde{Y}$. Consider the single-label problem and assume that the marginal distribution of label is a discrete uniform distribution over a finite set of size $M$, then:
\begin{align}
I(H;\tilde{Y})=\frac{1}{M}\sum_{i=1}^{M}D_{\text{KL}}(\mathbb{P}_{\tilde{y}_i}||\mathbb{\bar{P}})
\end{align}
where $\mathbb{\bar{P}}:=\frac{1}{M}\sum_{i=1}^{M}\mathbb{P}_{\tilde{y}_i}$, and $\mathbb{P}_{\tilde{y}_i}$ is the distribution of sample representation with the label $\tilde{y}_j$. In this paper, we use $y$ to denote the hypothetical label and $\tilde{y}$ to denote the ground-truth label. Estimating the precise mutual information value or finding a tractable upper bound is difficult. Since only the mutual information minimization is cared about, we circumvent the computation difficulty of mutual information by minimizing a tractable surrogate divergence measure. Here we employ Maximum Mean Discrepancy (MMD) since it can be efficiently estimated from samples using kernel trick.
The definition and our detailed implementation of MMD are given in appendix \ref{section:MMD}.

\textbf{Complete framework.}    The overall objective for the framework is obtained by summing up the proposed objective and regularizer. Details of the complete framework are given in Appendix \ref{algorithm}. We give an intuitive explanation of why better compression of mutual information improves adversarial robustness in Appendix \ref{section:intuitive}.

\section{Experiments}

The proposed framework is evaluated under various types of attacks in two threat modes on MNIST and Fashion-MNIST datasets. We compare our method with Alemi et al. \cite{Alemi} since both of the two approaches have information theory based regularizers and do not require adversarial training \cite{madry2017towards}.

Quantitative results are given under three attack methods: the FGSM attack \cite{goodfellow2014explaining} in $l_\infty$ norm, the PGD attack \cite{madry2017towards} in $l_\infty$ norm and the Carlini \& Wagner attack (C\&W) \cite{carlini2017towards} in $l_2$ norm. Each attack method is conducted in black-box mode and white-box mode.

We illustrate the decline of model accuracy with the increase of adversarial perturbation size and report the model accuracy under the attack of certain perturbation norms. Four models are evaluated: our proposed architecture with Equation \ref{eqn:dv} as objective (C-DV), our proposed architecture with Equation \ref{eqn:js} as objective (C-JS), and their corresponding regularized version (C-DV-IE, C-JS-IE) that are regularized by the proposed information elimination regularizer. More detailed experiment setup is given in the Appendix \ref{experiment details}.

\subsection{Results and Analysis}

\begin{table}
\centering
\small
\begin{tabular}{lccccccc}
\toprule
\textbf{Attack} & $\bm{\epsilon}$ & \textbf{CNN}  & \textbf{C-DV} & \textbf{C-JS} & \textbf{C-DV-IE} & \textbf{C-JS-IE} &
\textbf{Alemi et al.} \\
\midrule
Clean & 0 & 98.8 & 98.9 & \textbf{99.0} & 98.9 & \textbf{99.0} & \textbf{99.0} \\
\midrule
FGSM b & 0.15 & 65.6 & 84.7 & 88.0 & \textbf{91.1} & 89.8 & 83.6\\
 & 0.30 & 16.0 & 32.8 & 26.8 & \textbf{44.6} & 40.6 & 27.3 \\
PGD b & 0.15 & 53.7 & 84.1 & 85.0 & 87.2 & \textbf{87.6} & 74.3\\
 & 0.30 & 8.4 & 24.0 & 13.1 & \textbf{27.7} & 27.0 & 15.5\\
C\&W b & 2.00 & 63.4 & 76.0 & 74.4 & \textbf{79.3} & 74.8 & 66.7\\
\midrule
FGSM w & 0.15 & 40.8 & 54.0 & 75.6 & 58.7 & \textbf{82.8} & 50.7\\
 & 0.30 & 13.4 & 23.8 & 74.4 & 31.9 & \textbf{81.1} & 19.1\\
PGD w & 0.15 & 4.4 & 26.2 & 69.8 & 38.0 & \textbf{78.0} & 11.4\\
 & 0.30 & 0.0 & 4.8 & 29.7 & 7.4 & \textbf{50.0} & 1.0\\
C\&W w & 2.00 & 1.5 & 34.6 & 7.7 & \textbf{38.8} & 9.6 & 6.7\\ 
\bottomrule
\end{tabular}
\caption{Classification accuracy (\%) of different models under various type of attacks with different perturbation thresholds $\epsilon$ on MNIST. The perturbation norm for FGSM, PGD, C\&W is $l_\infty$, $l_\infty$, $l_2$ respectively. 'b' denotes black-box attack and 'w' denotes white-box attack.}
\label{table:1}
\end{table}

Results are illustrated in Figure \ref{fig:1} and Table \ref{table:1}, \ref{table:2} (see Figure \ref{fig:1} and Table \ref{table:2} in Appendix \ref{experiments results}). The proposed framework demonstrates all-round adversarial robustness improvements over the baseline CNN and Alemi et al. \cite{Alemi}, without loss of natural accuracy. Here the baseline model shares almost all other architecture similarities with our models.
We give analysis on MNIST for black-box attacks and white-box attacks separately.

\textbf{Black-box attack}  Black-box attack leverages the transferability of adversarial examples and invalidates the use of gradient masking, it is thus a consistent indicator of the model's robustness. 
While all our four models show significant robustness improvement in this mode on two datasets, the two models with DV-based objectives exceed the two with JS-based objectives, indicating the respective advantages of different objectives under our framework.
On MNIST, for $l_\infty$ attacks with $\epsilon=0.15$, FGSM successfully reduces the accuracy of the baseline model to 65.6\%, and PGD further reduces it to 53.7\%, whereas all our four models keep the accuracy above 84\%. With the increase of perturbation size, model accuracy rapidly drops. However, the proposed models consistently precede baseline models by a large margin. For FGSM attack with $\epsilon=0.3$, model C-DV-IE and C-JS-IE enjoys an accuracy boost of over 10\% in comparison with model C-DV and C-JS, demonstrating the effectiveness of the proposed regularizer. On Fashion-MNIST where samples are natural images, the baseline accuracy of all models drops, whereas the tendency stays the same as on MNIST.

Carlini \& Wagner (C\&W) attack is one of the most powerful attacks which solves a constrained optimization problem during attacking. The benefit provided by the regularizer under this attack shrinks to an inconspicuous but consistent improvement of about 3\% on MNIST. Nonetheless, models with the proposed framework still raise the accuracy from 63.4\% of baseline to 74.4\%-79.3\%, demonstrating the robustness improvement against $l_2$ norm based attacks.

\textbf{White-box attack}  Models with different objectives vary dramatically under white-box attacks. For $l_\infty$ attacks, the JS-divergence based objective potentially equips the model with gradient masking. As in Figure \ref{fig:1}, model C-JS and C-JS-IE surpass others by a marked gap. About 80\% of FGSM attacks fail to mislead C-JS and C-JS-IE under the experiment settings. All $l_\infty$ attacks under white-box mode are more effective in casting down the accuracy of C-DV, C-DV-IE and baseline model on both MNIST and Fashion-MNIST. However, C-DV and C-DV-IE still have higher accuracy in comparison with the baseline model. On MNIST, for FGSM with $\epsilon=0.15$, C-DV and C-DV-IE raise the baseline accuracy from 40.8\% to 54.0\% and 58.7\%, respectively. Noticeable robustness improvements also exist under PGD attacks, where accuracy is raised from 4.4\% to 26.2\% and 38.0\%. 

Situation changes when conducting $l_2$ attacks, where DV-based models show superior robustness. For C\&W attacks with $\epsilon=2$ on MNIST, baseline model has an accuracy of 1.5\%, C-JS and C-JS-IE have an accuracy of 7.7\% and 9.6\%, while C-DV and C-DV-IE keep the accuracy above 30\%.

Under both black-box and white-box attacks on the two datasets, our framework shows superiority in adversarial robustness over the baseline models, without loss of accuracy on clean samples. The extra cost in inference time is negligible ($<$5\%) since our GPU (8GB memory) loads all hypothesis vectors up and computes in parallel.

\section{Conclusion}

We present a novel classification framework with improved adversarial robustness. This framework includes an architecture that makes model inference conditioned on label hypothesis, a new class of objectives that offer different properties, and a regularizer that further improves robustness. We conduct experiments under various types of attacks and empirically demonstrate the effectiveness of our framework. Future work will be focused on theoretically explaining the effectiveness and exploring different properties of the objectives. Efficiency enhancement measures like hierarchical hypothesis or partial forward inference will also be investigated.


\medskip

\small

\bibliographystyle{unsrt}
\bibliography{reference.bib}

\newpage
\begin{appendices}

\section{Other Objective Representation Forms}\label{objectives}
\textbf{The Donsker-Varadhan representation}.  For KL-divergence, the Donsker-Varadhan representation has the form:
\begin{align}
D_{\text{KL}} (\mathbb{J} || \mathbb{M}) & \geq \sup_{T \in \mathcal{T}} \mathbb{E}_{\mathbb{J}}[T(x,y)] - \text{log}\mathbb{E}_{\mathbb{M}}[e^{T(x,y)}] \label{eqn:dv}
\end{align}

The bound is tight for $T^*(x,y)=1+\text{log}\frac{p(x, y)}{p(x)p(y)}$, which is identical to the optimal $T^*$ of KL-divergence in convex conjugate representation.

\textbf{The Energy-based Representation}. From the energy-based view in \cite{Poole}, we can derive a variational representation for KL-divergence:
\begin{align}
 D_{\text{KL}} (\mathbb{J} || \mathbb{M}) 
\geq \sup_{T \in \mathcal{T}} \mathbb{E}_{p(x,y)}[T(x,y)] - \mathbb{E}_{p(x)}[\text{log}\mathbb{E}_{p(y)}[e^{T(x,y)}]] \label{eqn:eb}
\end{align} 
We can see the connection between energy-based representation and the cross-entropy loss by sampling $Y$ uniformly in all categories to estimate the Expectation in Equation \ref{eqn:eb}:
\begin{align}
D_{\text{KL}} (\mathbb{J} || \mathbb{M})  \geq \max_{\theta \in \Theta} \mathbb{E}_{p(x)}[T_{\theta}(x,y_i) - \text{log}\sum_{j=1}^{M}{e^{T_{\theta}(x,y_j)}}] \label{eqn:ieb}
\end{align} 
where right hand side is exactly the log-softmax form in cross-entropy loss with conditional distribution $p(y|x)$ replaced by $T_\theta(x,y)$.

\section{Detailed implementation of Maximum Mean Discrepancy} \label{section:MMD}
\begin{figure}[t]
    \centering
    \includegraphics[width=0.6\textwidth]{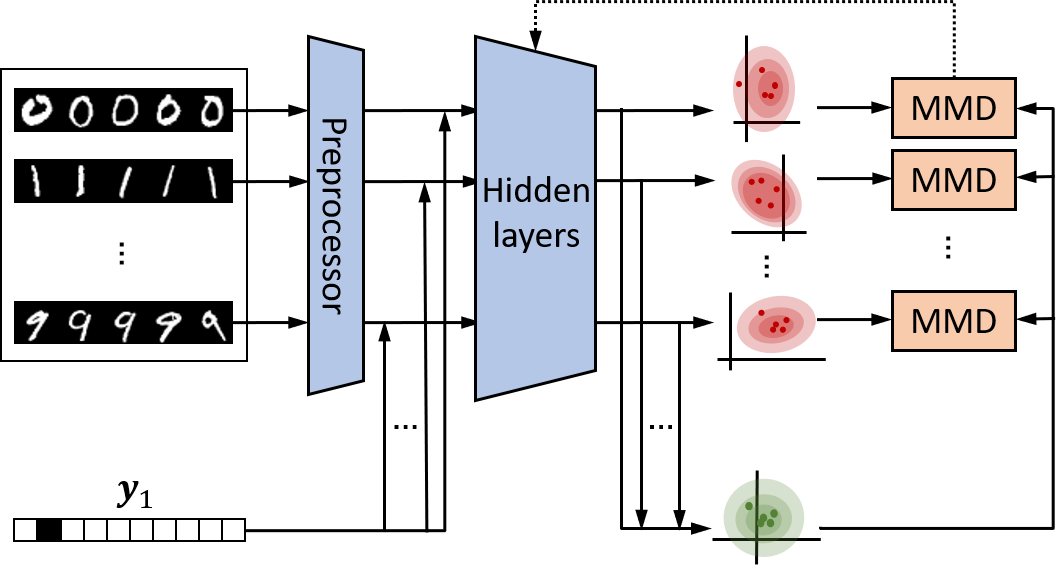}
    \caption{Illustration of the proposed regularizer. Intermediate outputs of specified hidden layers are used to calculate the regularization loss, with gradients backpropagated up to the layers where label hypotheses are concatenated. Samples within a batch are categorized according to their labels. For each one-hot label hypothesis that is concatenated (number '1' in this example), we obtain ten empirical distributions for each sample category (red) and an overall distribution (green). MMDs are then computed to obtain the regularization loss for this certain label hypothesis.}
    \label{fig:p2}
\end{figure}

Given two distribution $\mathbb{P}_{\tilde{y}}$ and $\mathbb{\bar{P}}$, the MMD is defined as the squared distance between their embeddings in reproducing kernel Hilbert space. Then with $n$ samples $h_i$ from  $\mathbb{P}_{\tilde{y}}$ and $m$ samples $\bar{h}_i$ from $\mathbb{\bar{P}}$ and a characteristic kernel $k$, we have its empirical estimate:
\begin{align}
\text{MMD}_k(\mathbb{P}_{\tilde{y}},\mathbb{\bar{P}}) &= ||\mu_{\mathbb{P}_{\tilde{y}}} - \mu_{\mathbb{\bar{P}}}||_\mathcal{H}^2 \\
&= \frac{1}{n^2}\sum_{i=1}^n \sum_{j=1}^n k(h_i,h_j) + \frac{1}{m^2}\sum_{i=1}^m \sum_{j=1}^m k(\bar{h}_i,\bar{h}_j) - \frac{2}{nm}\sum_{i=1}^n \sum_{j=1}^m k(h_i,\bar{h}_j) \notag
\end{align}
We use Gaussian kernel $k(x,x')=\text{exp}(-\frac{1}{2\sigma^2}||x-x'||^2)$ in our implementation. Figure \ref{fig:p2} illustrate the regularization imposed on specified hidden layers in MNIST handwritten digit recognition task. Note that only the loss of label hypothesis number '1' is displayed. We add up all ten losses for each label hypothesis to obtain the final regularization loss.

\section{Proposed Algorithm} \label{algorithm}

The complete training objective of the proposed framework is (adopting the variational representation of JS-divergence in Equation \ref{eqn:js} for example):
\begin{align}
\theta^*=&\argmin_{\theta}\frac{1}{B}\sum_{i=1}^{B}\text{log} \sigma (T_\theta(x_i,\tilde{y}_i))
- \frac{1}{BM}\sum_{i=1}^{B}\sum_{j=1}^{M}\text{log}(1- \sigma (T_\theta(x_i,y_j))) \notag\\
&+ \frac{\beta}{M^2}\sum_{i=1}^{M}\sum_{j=1}^{M}{\text{MMD}_k(}(\mathbb{P}_{\tilde{y}_j}^{(y_i)},\mathbb{\bar{P}}^{(y_i)}) \notag
\end{align}
where $B$ is the batch size, $M$ is the label category size, $\theta$ is the model parameter and $\beta$ is the regularization coefficient. Implementation details of the framework is given in Algorithm \ref{alg:1}. Since our framework advances label hypothesis input from output layer to the optional front layers, the growth ratio of inference complexity varies from zero (still input at output layer) to $O(M)$ (input at the frontmost layer). Efficiency enhancement measures like hierarchical hypothesis or partial forward inference can be leveraged.

The overall algorithm of this framework is given in Algorithm \ref{alg:1}

\begin{algorithm}
\caption{The complete framework}
\label{alg:1}
 {\bf Input:} 
Preprocessing layers $F_{conv}$, hidden layers (to be regularized) $F_{r}$ with parameter $\theta_r$, hidden layers (not to be regularized) $F_c$, model parameter $\theta$ (including $\theta_r$), batch size $B$, label category size $M$, regularizer coefficient $\beta$. \\
 {\bf Output:} 
$\theta^*$, optimized model parameter.

\begin{algorithmic}[1] 
\WHILE{$\theta$ not converge}
\STATE Sample a minibatch $\{x\}^B$ from the training set
\STATE Broadcast one-hot embeded label vectors:\\ $\{y\}^{B\times M} \gets \{y\}^{M}\times B \text{ times}$
\STATE Get preprocessor outputs: \\
$\{h_{conv}\}^B \gets F_{conv}(\{x\}^B)$ 
\STATE Broadcast preprocessor outputs:\\  $\{h_{conv}\}^{B\times M} \gets \{h_{conv}\}^B \times M \text{ times}$
\STATE Concatenate preprocessor outputs and label vectors:\\
$\{(h_{conv}:y)\}^{B\times M}\gets$ $(\{h_{conv}\}^{B\times M}\parallel \{y\}^{B\times M})$
\STATE Get hidden layer (to be regularized) outputs:\\
$\{h_r\}^{B\times M} \gets F_{r}(\{(h_{conv}:y)\}^{B\times M})$ 
\STATE Compute regularization loss $L_r$:\\ $L_r=\frac{\beta}{M^2}\sum_{y=1}^{M}\sum_{\tilde{y}=1}^{M}{\text{MMD}_k(}(\mathbb{P}_{\tilde{y}}^{(y)},\mathbb{\bar{P}}^{(y)})$
\STATE Get model outputs:\\
$\{T\}^{B\times M} = F_{c}(\{h_{r}\}^{B\times M})$ 
\STATE Compute objective loss $L_o$:\\
according to Equation \ref{eqn:js}, \ref{eqn:dv} or \ref{eqn:eb}
\STATE Backpropagate $L_o$ to update $\theta$
\STATE Backpropagate $L_r$ to update $\theta_r$
\ENDWHILE
\STATE \textbf{return} $\theta$
\end{algorithmic}
\end{algorithm}

\section{Why Compression Improves Adversarial Robustness}
\label{section:intuitive}

We give an intuitive explanation of why compression in terms of mutual information improves adversarial robustness. Note that minimizing mutual information between input samples and latent representations has already been shown helpful to standard generalization \cite{xu2017information,bassily2017learners} (via different ways such as stability). As for adversarial robustness, Pensia et al. \cite{pensia2019extracting} show that low Fisher information of input samples with respect to latent representations will lead to stable (in terms of KL-divergence) latent representation distributions after input adversarial perturbation, which further gives other model robustness properties. We intuitively extend these conclusions by utilizing the connection between mutual information and Fisher information.

Consider the problem formulation as the latent representation (denoted by random variable $Z$) is generated conditioned on input samples (denoted by random variable $X$). We first illustrate the connection between mutual information $I(X;Z)$ and the Fisher information $I_x$ (by adapting the result from \cite{brunel1998mutual}). Then we restate one of the results in \cite{pensia2019extracting}. Together they will give the desired intuitive explanation.

\begin{lemma}
\label{lemma:1}
Consider the Markov chain $X \xrightarrow{g} Z \xrightarrow{f} \hat{X}$, where $\hat{X}$ is, under the existence assumption, an unbiased efficient estimator with mean $X$ and variance $\frac{1}{I_x}$. Further assume $H(\hat{X})\rightarrow H(X)$ asymptotically. We have the following inequality between mutual information $I(X;Z)$ and Fisher information $I_x$:
\begin{align}
    I(X;Z) \geq \frac{1}{2} \E_X [\log I_x] + c
\end{align}
where $c$ is a constant irrelevant to $g$.
\end{lemma}

\textit{Proof.} Fisher information $I_x$ is defined as:
\begin{align}
    I_x = \int_\mathcal{Z} || \frac{\partial}{\partial x} \log g(z;x) ||^2 g(z;x)  dz
\end{align}
Mutual information $I(X;\hat{X})$ satisfies:
\begin{align}
    I(X;\hat{X}) = H(\hat{X}) - \int_\mathcal{X}H(\hat{X}|X=x)p(x)dx
\end{align}
By assumption, $\hat{X}$ is the unbiased efficient estimator with variance equal to the Cramér-Rao bound $\frac{1}{I_x}$. Given the variance, the entropy $H(\hat{X}|X=x)$ is maximized when the distribution $\PP(\hat{X}|X=x)$ is Gaussian (moment-entropy inequalities), that is $H(\hat{X}|X=x) \leq \frac{1}{2} \log (\frac{2\pi e}{I_x})$. Hence,
\begin{align}
    I(X;\hat{X}) \geq H(\hat{X}) - \int_\mathcal{X}\frac{1}{2} \log (\frac{2\pi e}{I_x})p(x)dx
\end{align}
Further by the Data-processing inequality $I(X;Z) \geq I(X;\hat{X})$ and the asymptotic assumption $H(\hat{X})\rightarrow H(X)$, we have:
\begin{align}
    I(X;Z) \geq H(X) - \int_\mathcal{X}\frac{1}{2} \log (\frac{2\pi e}{I_x})p(x)dx
\end{align}
Rearranging the inequality we complete the proof.

\begin{lemma}[Lemma 7 in \cite{pensia2019extracting}]
\label{lemma:2}
Consider the Markov chain $X \rightarrow Z$. Let $||u||_2=1$. Let $x+\varepsilon u$ be a small perturbation of $x$ in the (arbitrary) direction $u$. Then we have:
\begin{align}
    D(P_{Z|X=x+\varepsilon u} || P_{Z|X=x}) = \frac{\varepsilon^2}{2} I_x + o(\varepsilon^2)
\end{align}
where $P_{Z|X=x+\varepsilon u}$ is the distribution of $Z$ conditioned on perturbed $X=x+\varepsilon u$.
\end{lemma}

From Lemma \ref{lemma:1} and lemma \ref{lemma:2} we can see that under our problem formulation, low mutual information $I(X;Z)$ leads to low Fisher information $I_x$ logarithmically in expectation, under certain assumptions. Low Fisher information here further leads to stable latent representation distribution under small adversarial perturbation (and other following robustness properties as shown in \cite{pensia2019extracting}). This constitutes our intuitive explanation of why compression improves adversarial robustness.

\section{Related Works}

While the cause of adversarial examples is still an open problem with some synthetic pilot studies \cite{schmidt2018adversarially}, many papers are proposed to enhance the neural network's adversarial robustness \cite{akhtar2018threat}. Among those defending methods, adversarial training \cite{madry2017towards} is one of the most effective ways, however, it requires dramatically larger model capacity and computational resources to cover all types of samples and attacks and is therefore difficult to extend to large dataset (e.g. ImageNet). Some works provide certifiable robustness \cite{sinha2017certifying,wong2017provable,schott2018towards} which is a promising way, but currently, they either make strong assumptions or fail to scale up to large problems. Some defending approaches \cite{samangouei2018defense} fall into the gradient masking mechanism and are proved ineffective \cite{athalye2018obfuscated}. Some other papers utilize sparseness \cite{guo2018sparse} or information theory \cite{Alemi} to mildly improve the model's robustness. Despite those efforts, designing models that are robust to adversarial examples is still challenging.

To the best of our knowledge, we are the first to utilize the hypothesis-testing-like mechanism to improve classification adversarial robustness. A similar model architecture is proposed in \cite{hjelm2018learning}, in which they provide a novel unsupervised learning approach by maximizing the mutual information between samples and their representations. To estimate the mutual information, they combine samples and representations together and then train a critic to distinguish between true sample-representation pairs and the synthesized ones. Except for the different task goals, our architecture differs in that they do both estimation and maximization over the divergence of two changing distributions, whereas we only do estimation over the divergence of two fixed distributions and therefore our method enjoys the stability in training. 

The idea of improving adversarial robustness by constraining information flow is proposed in \cite{Alemi}. They introduce an information constraint on an intermediate layer, which is called information bottleneck, to constrain the mutual information between samples and their representations. Similar to the Variational Auto Encoder (VAE), they derive a variational upper bound for mutual information that can be viewed as a penalty of the KL-divergence between sample representation distribution and a prior distribution (normal distribution in their case). However, matching sample representation to a fixed prior is a strong assumption and may be too restrictive as a regularization. It can only be used after multiple layers' mapping to ensure that the model has enough capacity for the matching. Now that our goal is to constrain the mutual information between samples and their labels, we instead penalize the divergence between sample representations and a data-dependent prior distribution, thus making the regularizer more efficient and task-oriented.

Some similarities also exist between our work and \cite{schott2018towards}. To arrive at an adversarial robust model on MNIST that has practical nonempty robust bound, they parallel multiple VAEs, each for one label category, to approximate the evidence probability.  During inference, for one input sample, each VAE gives an optimized evidence probability followed by a trained weight, and the prediction is made by finding the largest output value. Their work gives insight on training separate binary classifiers and keeps each one of them ignorant of label information. Despite the novel idea and convincing result, their model suffers from several disadvantages, such as unable to scale up to a large dataset, and the optimization-based inference is rather slow. Besides, the robust bound they derive is valid only when the optimization gives a high quality solution. In contrast, our framework is easy to scale and does not need optimization during inference.

\section{Experiment Details}\label{experiment details}
\textbf{Implementation details} The basic architecture includes two convolutional layers with 32 and 64 filters of size 4 $\times$ 4 and stride 2, and three fully-connected layers of size 512, 256, 128, respectively. We use batch-normalization after convolutional layers and ReLU activation after each layer except the last one. For baseline CNN, we add a fully-connected layer of size 10 at the end with softmax activation. For model C-DV and C-JS, we concatenate one-hot label vectors to the output of the second convolutional layer and add a fully-connected layer of size 1 at last without activation. For model C-DV-IE and C-JS-IE, we use the output of the second fully-connected layer for MMD computation and backpropagate the obtained gradient to update weights of the first and second fully-connected layers. We reimplemented the framework in \cite{Alemi} by concatenating the information bottleneck layer with stochastic encoders to the middle layer of our baseline model. For the substitute model in black-box attacks, we train a LeNet-5-like CNN model on the full training set. Random start is enabled for PGD attack. We set max perturbation size to 1 for FGSM attack, and the perturbation size will be recorded as infinity if an attack fails to find the adversarial example for a test image.

In black-box mode, attackers have no knowledge about the defense model, a substitute is therefore needed to guide the attack. While in the white-box mode, attackers have full knowledge about the defense model, including the architecture and the weights, thus gradients can be computed precisely in this case. Foolbox \cite{rauber2017foolbox} is applied as the implementation of different attacks. 

\textbf{Hyperparameters}  We use Adam optimizer with learning rate 0.001. Each model is trained for 200 epochs with batch size 250. In our implementation, the estimated objective is not invariant with batch size, so we ensure a fixed batch size within each epoch.  We set regularization coefficient $\beta=0.001$ for our models and for \cite{Alemi}. We set $\sigma^2=1$ for Gaussian kernel.

\textbf{The concept of Gradient Masking} Here we introduce the term "gradient masking" \cite{papernot2017practical}. A defending model is said to use a gradient masking strategy if it provides useless gradients to the attacker, such as gradients with massive noise, gradients with large condition number, or misleading gradients.
Many defending approaches that behave well only on white-box attacks actually use the gradient masking strategy. It is known to be an incomplete defense strategy since attackers can easily circumvent it by estimating the correct gradients.
 
\newpage
 \section{Supplementary Experimental Results}\label{experiments results}
\begin{table}[h]
\centering
\small
\begin{tabular}{lccccccc}
\toprule
\textbf{Attack} & $\bm{\epsilon}$ & \textbf{CNN}  & \textbf{C-DV} & \textbf{C-JS} & \textbf{C-DV-IE} & \textbf{C-JS-IE} &
\textbf{Alemi et al.} \\
\midrule
Clean & 0 & 91.7 & 91.3 & 92.0 & 90.4 & 91.0 & \textbf{92.3} \\
\midrule
FGSM b & 0.15 & 18.2 & 52.3 & 47.9 & \textbf{54.9} & 51.3 & 29.3 \\
 & 0.30 & 9.6 & 24.2 & 19.8 & \textbf{28.8} & 24.7 & 15.3 \\
PGD b & 0.15 & 12.4 & \textbf{48.6} & 45.7 & 48.0 & 47.6 & 24.6 \\
 & 0.30 & 4.8 & 15.3 & 11.8 & \textbf{15.9} & 14.5 & 6.6 \\
C\&W b & 2.00 & 14.9 & 20.5 & 19.1 & \textbf{24.7} & 21.6 & 18.9 \\
\midrule
FGSM w & 0.15 & 14.4 & 32.6 & 43.5 & 33.8 & \textbf{49.3} & 27.9 \\
 & 0.30 & 7.9 & 15.2 & 39.7 & 18.9 & \textbf{44.0} & 12.5 \\
PGD w & 0.15 & 3.6 & 14.0 & 38.5 & 17.0 & \textbf{41.2} & 7.7 \\
 & 0.30 & 0.0 & 2.6 & 23.4 & 5.7 & \textbf{33.5} & 0.0 \\
C\&W w & 2.00 & 0.0 & 8.4 & 2.9 & \textbf{11.7} & 3.5 & 0.0 \\
\bottomrule
\end{tabular}
\caption{Classification accuracy (\%) of different models under various type of attacks with different perturbation thresholds $\epsilon$ on Fashion-MNIST. The perturbation norm for FGSM, PGD, C\&W is $l_\infty$, $l_\infty$, $l_2$ respectively. 'b' denotes black-box attack and 'w' denotes white-box attack. }
\label{table:2}
\end{table}

\begin{figure*}[h]
    \centering
    \subfloat[FGSM b ($l_\infty$)]{{\includegraphics[width=0.24\textwidth]{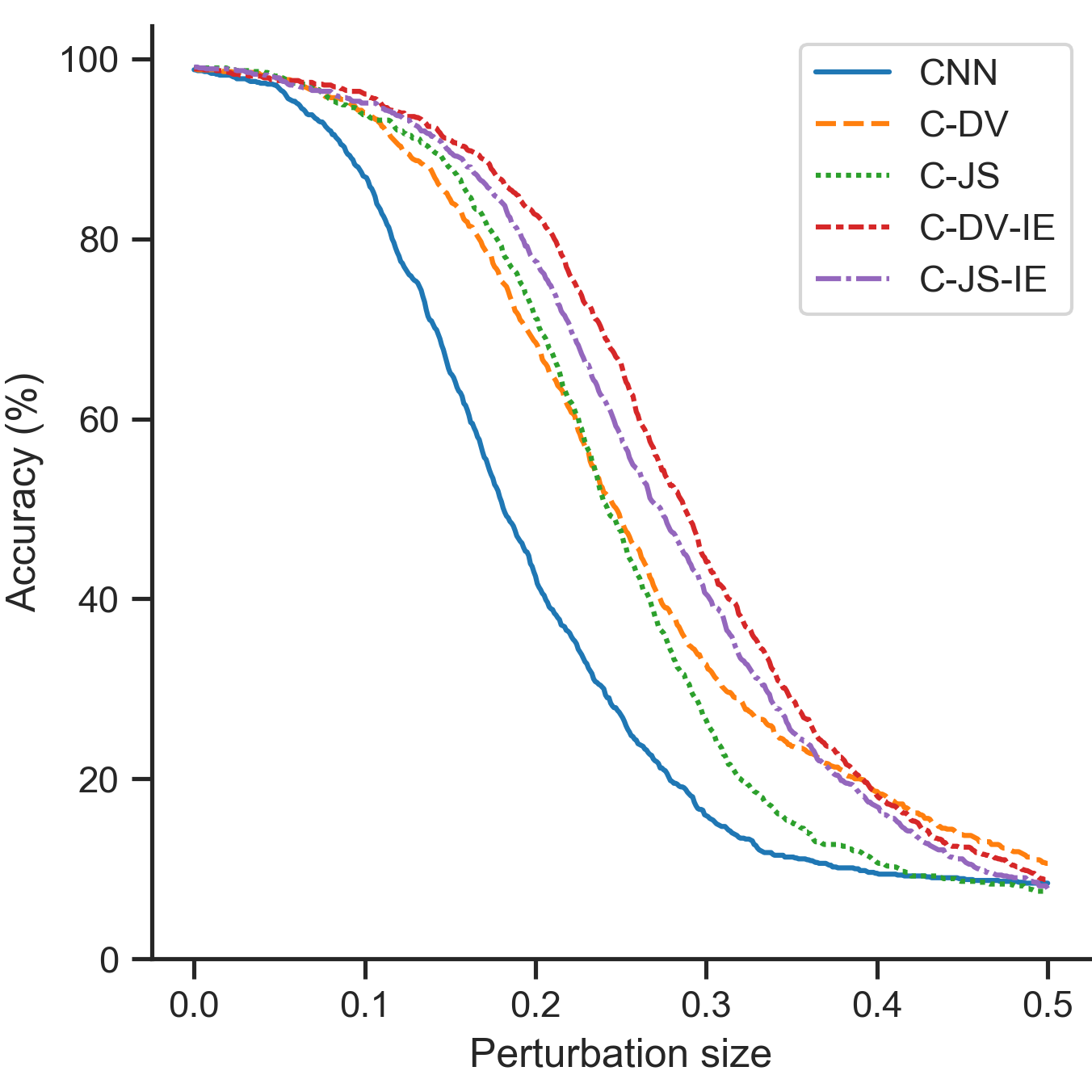}}}%
    \qquad
    \subfloat[PGD b ($l_\infty$)]{{\includegraphics[width=0.24\textwidth]{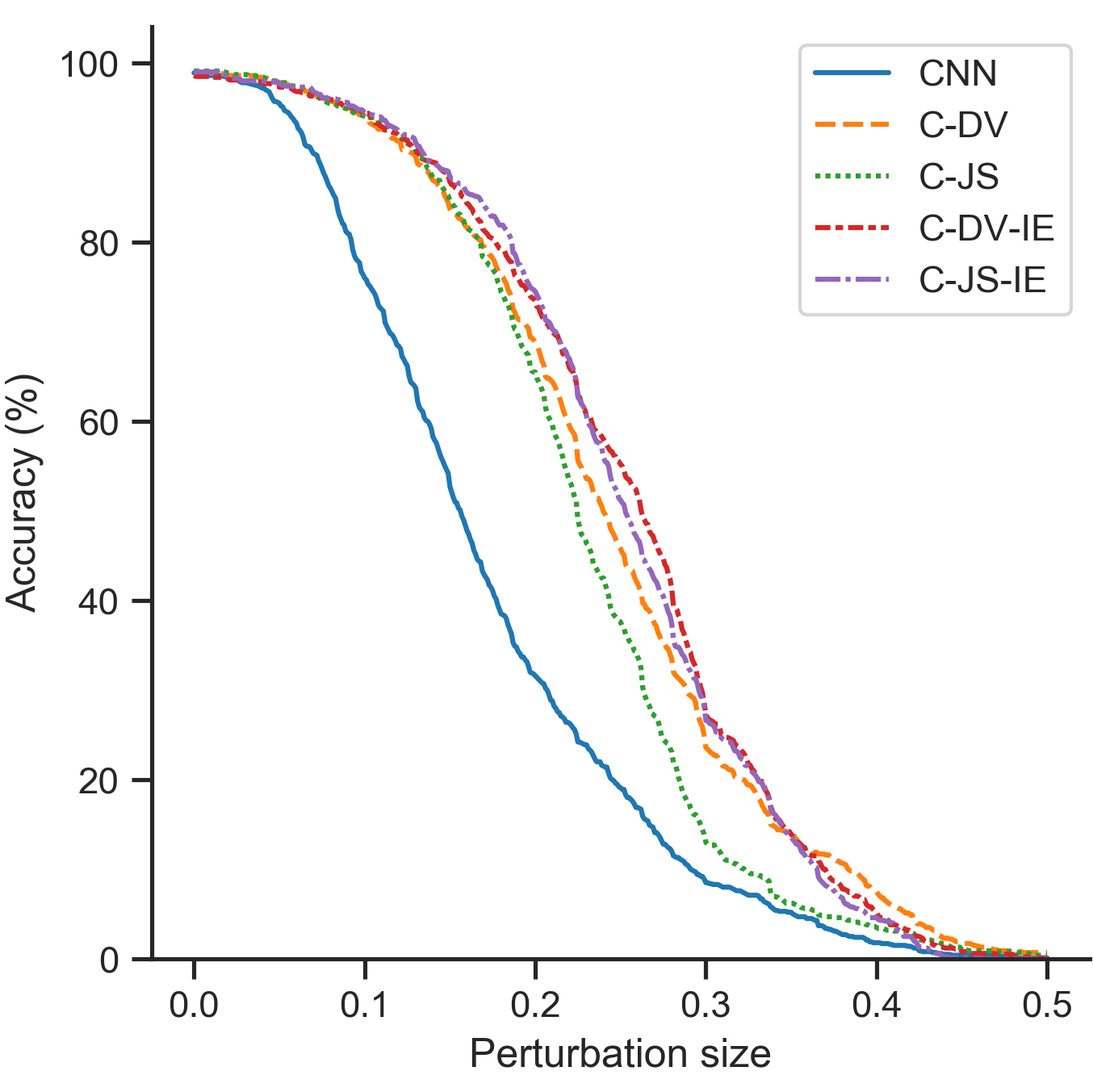}}}%
    \qquad
    \subfloat[C\&W b ($l_2$)]{{\includegraphics[width=0.24\textwidth]{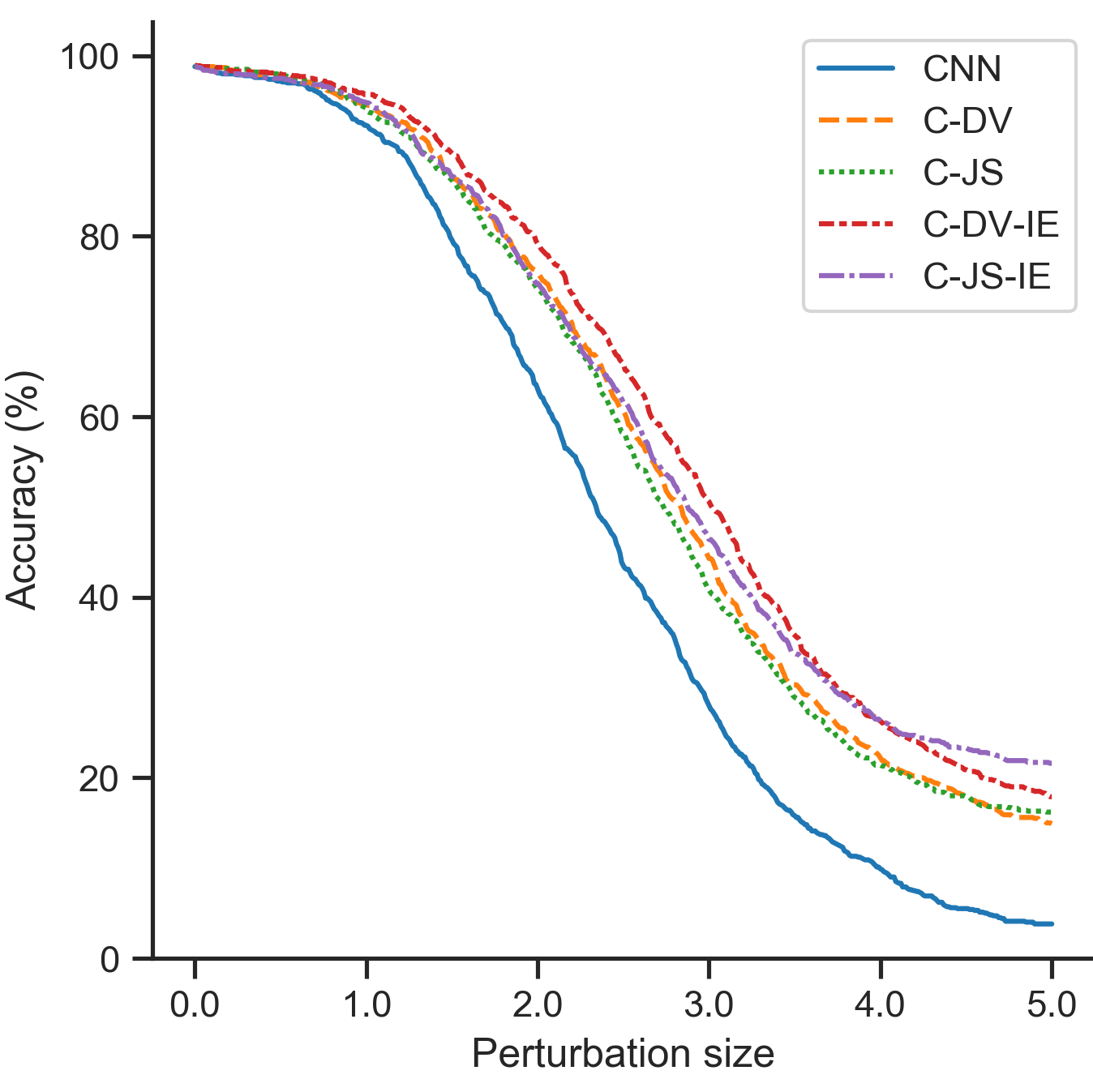}}} \\
    \subfloat[FGSM w ($l_\infty$)]{{\includegraphics[width=0.24\textwidth]{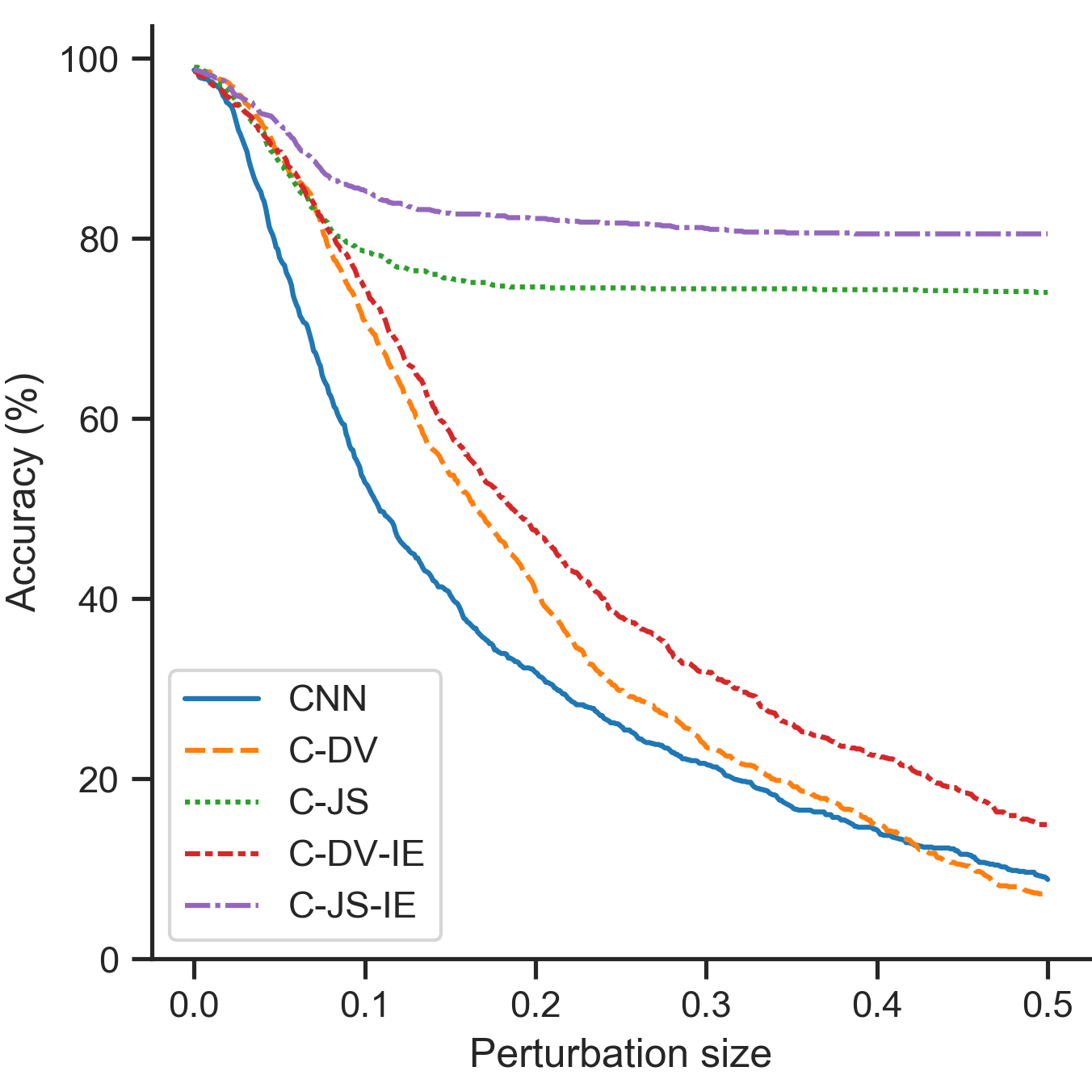}}}%
    \qquad
    \subfloat[PGD w ($l_\infty$)]{{\includegraphics[width=0.24\textwidth]{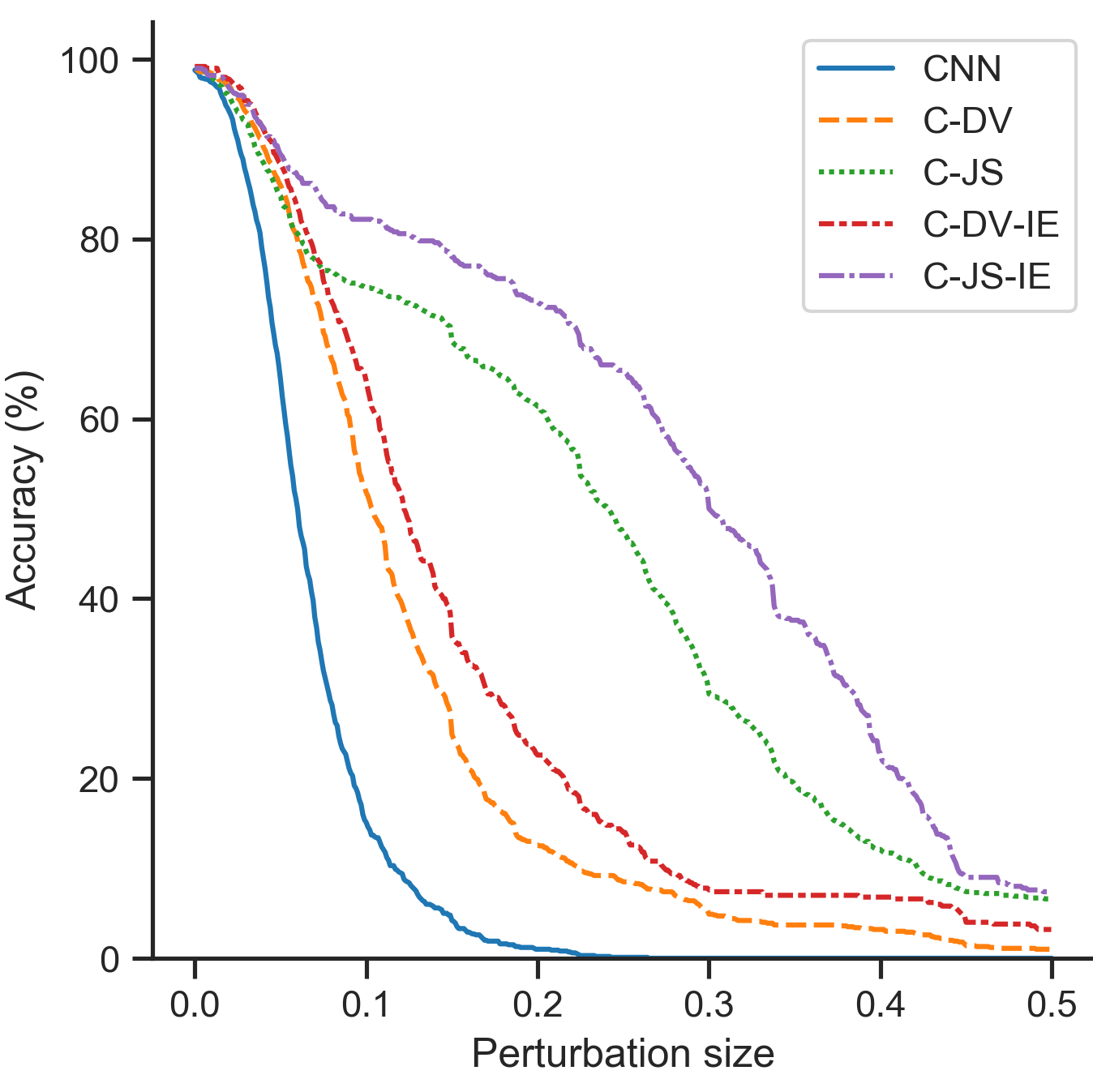}}}%
    \qquad
    \subfloat[C\&W w ($l_2$)]{{\includegraphics[width=0.24\textwidth]{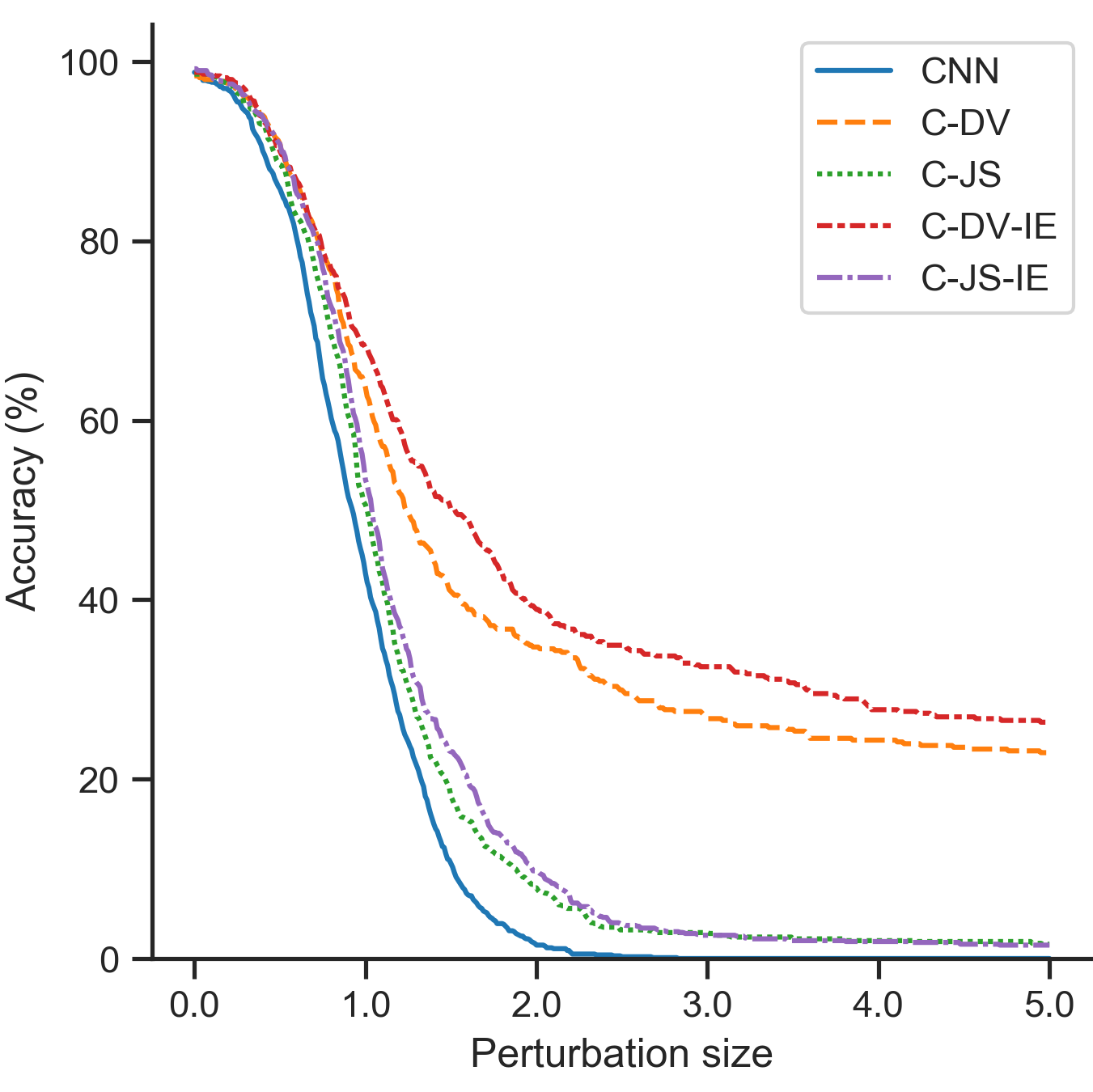}}}%
    \caption{Classification accuracy on MNIST decline with the increase of adversarial perturbation size. 'b' denotes black-box attacks and 'w' denotes white-box attacks.}%
    \label{fig:1}%
\end{figure*}



\end{appendices}

\end{document}